\documentclass[sigconf,natbib=true]{acmart}
\AtBeginDocument{%
  \providecommand\BibTeX{{%
    \normalfont B\kern-0.5em{\scshape i\kern-0.25em b}\kern-0.8em\TeX}}}

\copyrightyear{2022}
\acmYear{2022}
\setcopyright{acmlicensed}
\acmConference[SIGIR '22]{Proceedings of the 45th International ACM SIGIR Conference on Research and Development in Information Retrieval}{July 11--15, 2022}{Madrid, Spain}
\acmBooktitle{Proceedings of the 45th International ACM SIGIR Conference on Research and Development in Information Retrieval (SIGIR '22), July 11--15, 2022, Madrid, Spain}
\acmPrice{15.00}
\acmDOI{10.1145/3477495.3531824}
\acmISBN{978-1-4503-8732-3/22/07}

\usepackage{microtype}
\usepackage{graphicx}

\usepackage{amssymb}
\usepackage{amsthm}
\usepackage{amsmath}
\usepackage{verbatim}
\usepackage{colortbl}
\usepackage{multirow}
\definecolor{bestcolor}{RGB}{198,224,180}
\definecolor{secondcolor}{RGB}{221,235,247}
\definecolor{headercolor}{RGB}{242,242,242}

\begin{document}

\fancyhead{}
\title{Expression Syntax Information Bottleneck\\ for Math Word Problems}


\author{Jing Xiong}
\email{xiongj69@mail2.sysu.edu.cn}
\orcid{0000-0003-2986-6978}
\affiliation{%
  \institution{Sun Yat-sen University}
  \city{Shenzhen}
  \state{Guangdong}
  \country{China}
  \postcode{510275}
}
\author{Chengming Li}
\authornote{Corresponding authors.}
\orcid{0000-0002-4592-3875}
\email{lichengming@mail.sysu.edu.cn}
\affiliation{%
  \institution{Sun Yat-sen University}
  \city{Shenzhen}
  \state{Guangdong}
  \country{China}
  \postcode{510275}
}

\author{Min Yang}
\authornotemark[1]
\email{min.yang@siat.ac.cn}
\affiliation{%
  \institution{SIAT, Chinese Academy of Sciences}
  \city{Shenzhen}
  \state{Guangdong}
  \country{China}
  \postcode{518055}
}
\author{Xiping Hu}
\email{huxiping@mail.sysu.edu.cn}
\affiliation{%
  \institution{Sun Yat-sen University}
  \city{Shenzhen}
  \state{Guangdong}
  \country{China}
  \postcode{510275}
}
\author{Bin Hu}
\email{bh@lzu.edu.cn}
\affiliation{%
  \institution{Lanzhou University}
  \city{Lanzhou}
  \state{Gansu}
  \country{China}
  \postcode{730000}
}

\begin{comment}
\author{Ben Trovato}
\authornote{Both authors contributed equally to this research.}
\email{trovato@corporation.com}
\orcid{1234-5678-9012}
\author{G.K.M. Tobin}
\authornotemark[1]
\email{webmaster@marysville-ohio.com}
\affiliation{%
  \institution{Institute for Clarity in Documentation}
  \streetaddress{P.O. Box 1212}
  \city{Dublin}
  \state{Ohio}
  \country{USA}
  \postcode{43017-6221}
}

\author{Lars Th{\o}rv{\"a}ld}
\affiliation{%
  \institution{The Th{\o}rv{\"a}ld Group}
  \streetaddress{1 Th{\o}rv{\"a}ld Circle}
  \city{Hekla}
  \country{Iceland}}
\email{larst@affiliation.org}
\end{comment}


\begin{abstract}
Math Word Problems (MWP) aims to automatically solve mathematical questions given in texts. Previous studies tend to design complex models to capture additional information in the original text so as to enable the model to gain more comprehensive features. In this paper, we turn our attention in the opposite direction, and work on how to discard redundant features containing spurious correlations for MWP. To this end, we design an \textbf{E}xpression \textbf{S}yntax \textbf{I}nformation \textbf{B}ottleneck method for MWP (called \emph{ESIB}) based on variational information bottleneck, which extracts essential features of expression syntax tree while filtering latent-specific redundancy containing syntax-irrelevant features. The key idea of \emph{ESIB} is to encourage multiple models to predict the same expression syntax tree for different problem representations of the same problem by mutual learning so as to capture consistent information of expression syntax tree and discard latent-specific redundancy. To improve the generalization ability of the model and generate more diverse expressions, we Pdesign a self-distillation loss to encourage the model to rely more on the expression syntax information in the latent space. Experimental results on two large-scale benchmarks show that our model not only achieves state-of-the-art results but also generates more diverse solutions. The code is available.\footnote{https://github.com/menik1126/math\_ESIB}
\end{abstract}




\maketitle

\section{Introduction}
Math Word Problems (MWP) is challenging and draws much attention from researchers in the field of natural language processing \cite{bobrow1964natural, xiong2023dq} and information retrieval (e.g., mathematical understanding) \cite{huang2020neural,zanibbi2021arqmath}. MWP aims to automatically answer mathematical questions given in a natural language, which requires the model not only understand what facts are presented in a text, but also possess the reasoning capability to answer the mathematical question. Table \ref{tab:case0} shows three examples of MWP with three mathematical problems and their solution expressions with answer.

Inspired by the success of deep learning \cite{DBLP:journals/corr/abs-2302-09019,DBLP:conf/icml/0005SLW0X22}, attention-based Seq2Seq models \cite{bahdanau2014neural} have been dominated in MWP \cite{wang2019template, wang2018translating, wang2018mathdqn}, which bring the state-of-the-art to a new level.
The key idea is to use an encoder to learn representations of problem text and employ a decoder to generate the corresponding solution expression and answer.
Subsequently, several studies propose sequence-to-tree models, which explore the tree structure information presented in the text and improve the generation of solution expressions \cite{xie2019goal, zhang2020graph, wu2021math}.

However, the previous MWP methods appear to rely on spurious correlations between the {shallow heuristics in problem} and solution expression \cite{patel2021nlp}. For example,
%tend to find some shortcuts to solve problems.
as shown in Table \ref{tab:case0},
previous models may associate Problem 1 and Problem 2 with the mathematical formula ``x$\times$y$\div$z'', since these two problems have similar semantic patterns, e.g., \textit{calculating the speed}.
Based on this association, the models could generate wrong solution expression for Problem 3 which has similar semantic problem expression {like the text segment "place A to place B" in Problems 1-2.} In particular, the models that learn spurious correlations are more likely to generate wrong solution expression ``$220\times25\%\div30\%$'', rather than ``$220\div(25\%+30\%)$'' for Problem 3.
We define such a false association as spurious correlation.

Some recent studies have revealed that MWP solvers relying on spurious correlations could achieve high accuracy \cite{patel2021nlp,kumar2021adversarial}. These models can even compute correct answers without paying attention to the question part  in the problem such as the text segment ``\textit{how many kilometers is the total length of the two places?}'' in Problem 3 \textit{calculating the distance}. In addition, the solution expression is sensitive to the perturbed latent {representations} \cite{liangsolving}, since the semantically similar mathematical problems, even with totally different solution expressions and {questions}, can be encoded closely in the latent space. {We believe it is an evidence that redundant information containing spurious correlations is encoded in the latent representation.}
%MWP with different math expressions but similar mathematical problems are encoded closely in the latent space.
Therefore, it is necessary to alleviate the spurious correlations by {compressing the latent representations for math expressions while filtering latent-specific redundancy}.

To solve the above challenges, we design a \textbf{E}xpression \textbf{S}yntax \textbf{I}nformation \textbf{B}ottleneck method for MWP based on variational information bottleneck (VIB) \cite{alemi2016deep}, which aims to discard redundant information containing spurious correlations \cite{fischer2020conditional,mukherjee2020ccmi}.  Our key idea is to encourage multiple models to predict the same math expression with different problem representations of the same problem so as to capture consistent information about expression syntax tree in expressions and discard latent-specific redundancy
containing syntax-irrelevant information. In addition, we leverage mutual learning \cite{zhang2018deep} for learning variational information bottleneck, which can effectively reduce the latent-specific redundancy. Inspired by the observation that there are usually multiple solutions to solve a problem, we also design a self-distillation loss which encourages the decoder to rely on the syntax information in latent space, enabling the model to generate diverse solutions.

We summarize our main contributions as follows.
(1) We are the first to reduce spurious correlations so as to improve the performance of MWP.
(2) We propose a novel expression syntax information bottleneck method for MWP, which extracts essential syntax information of {math expression} and filters redundant information containing spurious correlations.
(3) We design a self-distillation loss to encourage the model to generate more diverse solution expressions.
(4) Extensive experiments on two benchmark datasets show that our model outperforms the strong baselines in a noticeable margin.

\section{Related Work}

\subsection{Math Word Problem Solving}
Math word problem (MWP) solving has been studied for decades. Early work \cite{bobrow1964natural} attempted to solve algebra word problems using rule-based approaches with hand-crafted features. These traditional methods rely on predefined templates and rules to map natural language problems to mathematical expressions, which limits their generalization capability.

With the development of deep learning, neural network-based methods have achieved significant progress in MWP solving. Wang et al. \cite{wang2017deep} first introduced a Seq2Seq model with attention mechanism to generate mathematical expressions from problem texts. Subsequently, Wang et al. \cite{wang2018translating} proposed an equation normalization method to reduce the diversity of equivalent equations. To leverage the tree structure of mathematical expressions, Xie and Sun \cite{xie2019goal} proposed a goal-driven tree-structured approach (GTS) that generates expression trees in a top-down manner. Zhang et al. \cite{zhang2020graph} further introduced Graph2Tree, which uses graph neural networks to capture the relationships between quantities in the problem text.

Recent studies have explored various techniques to enhance MWP solving. Template-based methods \cite{wang2019template} combine neural networks with predefined templates to improve accuracy. Multi-encoder and multi-decoder architectures \cite{shen2020solving} have been proposed to capture diverse representations. In addition, knowledge-aware approaches \cite{wu2020knowledge, wu2021math} incorporate external knowledge to improve reasoning capability, and teacher-student frameworks \cite{zhang2020teacher} utilize knowledge distillation to enhance performance.

However, recent studies \cite{patel2021nlp, kumar2021adversarial} have revealed that existing MWP solvers tend to rely on spurious correlations between surface patterns and solution expressions, rather than truly understanding the mathematical reasoning process. This motivates us to design methods that can effectively filter out such spurious correlations.

\subsection{Information Bottleneck}
The Information Bottleneck (IB) principle was first introduced by Tishby et al. \cite{tishby2000information}, which provides a theoretical framework for learning compressed representations that preserve task-relevant information while discarding irrelevant details. The core idea is to find a representation $Z$ that maximizes the mutual information $I(Z; Y)$ with the target $Y$ while minimizing the mutual information $I(X; Z)$ with the input $X$.

\paragraph{Variational Information Bottleneck}
Alemi et al. \cite{alemi2016deep} proposed the Deep Variational Information Bottleneck (VIB), which enables the application of IB principle to deep neural networks through variational inference. By introducing a variational approximation to the intractable IB objective, VIB provides a tractable lower bound that can be optimized using standard backpropagation. The key insight is to model the encoder as a stochastic mapping $p(z|x)$ parameterized by a neural network, enabling end-to-end training. VIB has been successfully applied to various tasks, including image classification and representation learning, demonstrating improved robustness and generalization.

Achille and Soatto \cite{achille2018information} introduced Information Dropout, which establishes a theoretical connection between dropout regularization and the information bottleneck principle. They showed that multiplicative noise injection in neural networks can be interpreted as minimizing the mutual information between the input and the learned representation, providing a principled understanding of dropout's regularization effect.

\paragraph{Extensions and Variants}
Several extensions to VIB have been proposed to address its limitations and expand its applicability. Kolchinsky et al. \cite{kolchinsky2019nonlinear} introduced the Nonlinear Information Bottleneck, which relaxes the assumption of Gaussian distributions and provides tighter bounds on the IB objective. This extension enables more flexible representation learning for complex data distributions.

Fischer \cite{fischer2020conditional} extended the IB framework to conditional settings, proposing the Conditional Entropy Bottleneck (CEB) that considers task-specific compression. CEB reformulates the IB objective to focus on conditional entropy, leading to representations that are more directly optimized for the downstream task.

Federici et al. \cite{federici2020learning} proposed Multi-view Information Bottleneck (MIB) for learning robust representations by encouraging consistency across different views of the same data. MIB decomposes the representation into view-specific and view-invariant components, enabling the extraction of shared semantic information while filtering view-specific noise.

\paragraph{Applications in NLP}
The information bottleneck principle has also been applied to natural language processing tasks. Mahabadi et al. \cite{mahabadi2021variational} proposed Variational Information Bottleneck for semi-supervised text classification, demonstrating that VIB can effectively leverage unlabeled data by learning compressed representations. Li and Eisner \cite{li2019specializing} applied VIB to word embeddings, showing that task-specific compression can improve performance on downstream NLP tasks by removing irrelevant semantic information.

These methods demonstrate the effectiveness of information bottleneck in extracting essential features while filtering redundant information, which motivates our application of VIB to math word problem solving for reducing spurious correlations.

\subsection{Mutual Learning and Knowledge Distillation}
Knowledge distillation \cite{hinton2015distilling} is a technique where a smaller student model learns from a larger teacher model by mimicking its output distributions. This approach has been widely adopted for model compression and transfer learning. Zhang et al. \cite{zhang2018deep} proposed Deep Mutual Learning (DML), where multiple student networks learn collaboratively and teach each other throughout the training process, without requiring a pre-trained teacher model.

In the context of MWP solving, teacher-student frameworks have been explored to improve model performance. Liang and Zhang \cite{liangsolving} proposed a teacher supervision method for solving math word problems. Zhang et al. \cite{zhang2020teacher} introduced a teacher-student network with multiple decoders to generate diverse solution expressions.

Our work differs from previous approaches by combining the information bottleneck principle with mutual learning for MWP solving. We leverage mutual learning to identify and filter latent-specific redundancy containing spurious correlations, while preserving essential syntax information of mathematical expressions.

\begin{table}
\caption{Three examples of MWP. }
\label{tab:case0}
\small
\begin{tabular}{|p{8cm}|}%lc}
\hline
{\textbf{Problem 1:} From place A to place B, if a bicycle travels 16 kilometers per hour, it can be reached in 4 hours. If it only takes 2 hours by car, how many kilometers per hour does the car travel?}\\
\hline
{\textbf{Solution Expression 1:} 16$\times$4$\div$2 \quad \textbf{Answer: }32}\\
\hline
{\textbf{Problem 2:} Uncle Jack drove from place A to place B, and it took 6 hours to arrive at the speed of 70 kilometers per hour. When I returned, I accelerated the speed due to the task. It only took 4 hours to return to the first place. What was the speed when I returned?}\\
\hline
{\textbf{Solution Expression 2:} 70$\times$6$\div$4 \quad \textbf{Answer: }105}\\
\hline
{\textbf{Problem 3:} A car travels 25\% of the whole journey from place A to place B in the first hour, 30\% of the whole journey in the second hour, a total of 220 kilometers in two hours, how many kilometers is the total length of the two places?}\\
\hline
{\textbf{Solution Expression 3:} $220\div(25\%+30\%)$ \quad \textbf{Answer: }400}\\
{\textbf{Wrong Solution Expression 3:} $220\times25\%\div30\%$}\\
\hline
\end{tabular}
\end{table}

\section{Methodology}

A math word problems (MWP) can be denoted by a projection $F: x \mapsto z \mapsto y$, where ${x}=\{w_1,w_2,\dots,w_m\}$ is the problem sequence with $m$ words, $z$ is the compressed representation of the original problem $x$ and $y=\{o_1,o_2,\dots,o_n\}$ is the solution expression of the problem with $n$ words. The goal of MWP is to establish a model $F$ which generates a correct solution expression $y$ and calculates the correct answer for the problem $x$.

\begin{figure}
     \centering
     \includegraphics[width=0.45\textwidth]{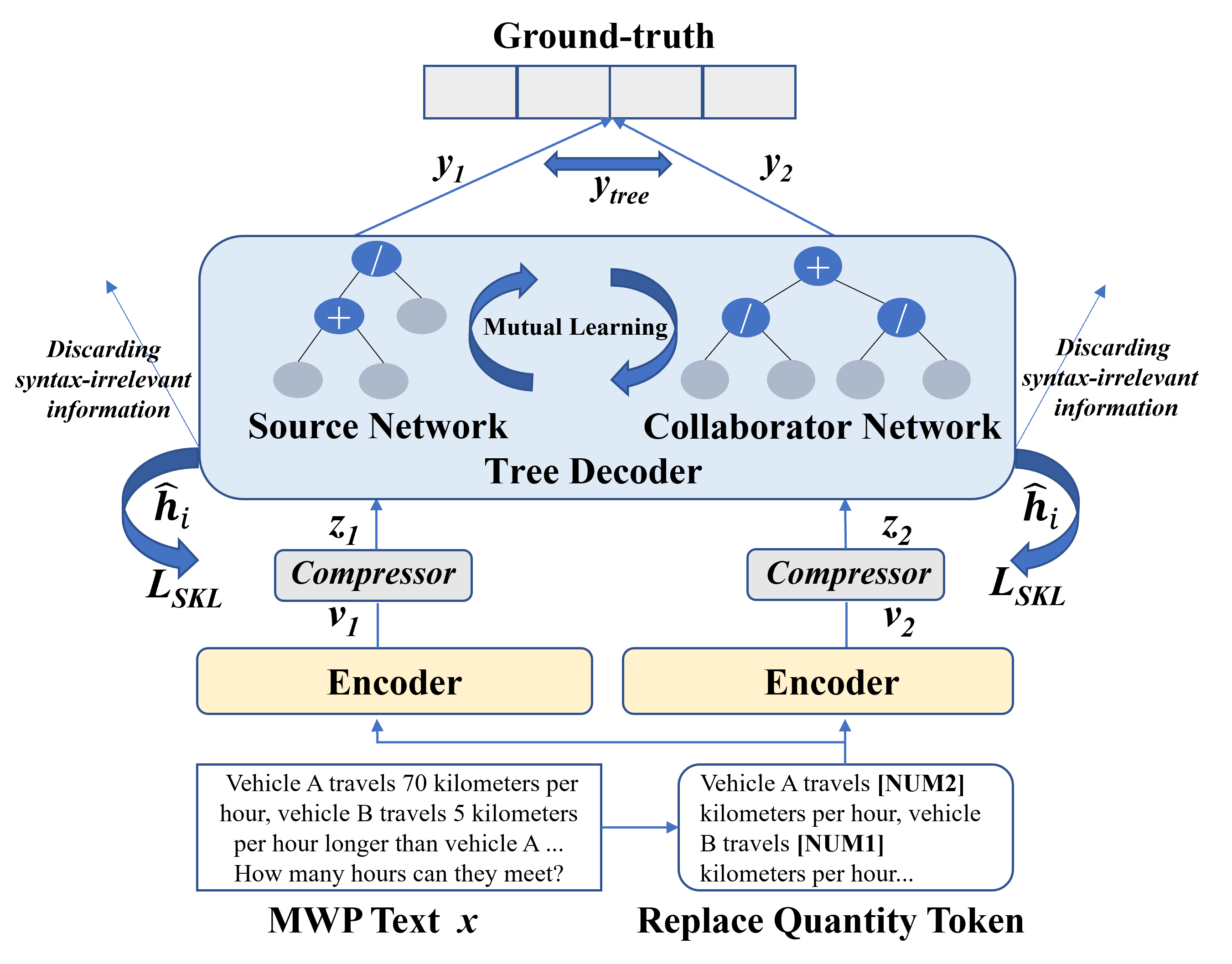}
     \caption{Overview of the proposed method ESIB. } \label{fig2}
\end{figure}

As illustrated in Figure \ref{fig2}, the proposed \emph{ESIB} is composed of a source network (denoted as SN) and a collaborator network (denoted as CN).  The two networks are optimised collaboratively and capture consistent information across different problem representations throughout the training process. We use the deep variational information bottleneck (VIB) framework as the backbone of both SN and CN. The VIB aims to generate problem representation $z \in \mathbb{R}^{s}$ by compressing and discarding redundant information in latent representation $v \in \mathbb{R}^{d}$ without reducing essential information related to the target $y$, where $d$ denotes hidden size of decoder and $s$ denotes the dimension of problem representation. So the above projection $F$ can be rewritten as $F: x \mapsto v \mapsto z \mapsto y$, where $v$ denotes latent representation and $z$ denotes problem representation sampled from $e^{\mu}(v) \sim N\left(e^{\mu}(v), e^{\sigma}(v)\right)$ ($e$ denotes a dense layer).

\subsection{Encoder-Compressor-Decoder Architecture}
%We briefly introduce the encoder, the compressor and the decoder.

\paragraph{Encoder}
We adopt the RoBERTa model \cite{liu2019roberta} as our encoder. We pass the problem sequence $x$ into the RoBERTa model and obtain latent representation $v$ ({[CLS] vector output by the pre-trained language model (PLM) and its dimension is converted from 768 to $d$ by a dense layer}).  In order to model the relationship between the quantities in the PLM, we set up a learnable quantity embedding matrix $\emph{Q}_t=\{q_1,q_2,\dots,q_n\}$, similar to the learnable position embedding in BERT \cite{devlin2018bert}. When passing the sequence $x$ into the encoder, we replace the each quantity in the sequence $x$ (i.e., the numbers in the problem) with a embedding $q_i \in \emph{Q}_t$.

\paragraph{Compressor}
Our model takes the encoder output $v$ and feeds it into a variational information bottleneck \cite{alemi2016deep} module which outputs a sampled vector $z$. This part removes redundant information by optimizing a variational upper bound $\mathcal{V}_{IB}$. We will detail how to optimize $\mathcal{V}_{IB}$ in Section 2.2.

\paragraph{Decoder}
Our decoder follows the GTS model \cite{xie2019goal}.
We use the sampled representation $z$ to initialize the initial state of the decoder and the recursive operation of the decoder to construct $y$ by the order of pre-order traversal. First, the root node $q_{\rm root}$ (middle operator part) is first generated. Then, we generate the left child node $q_{l}$. Finally, we generate the right child node $q_{r}$. This process has been iterated until the leaf nodes are generated. The attention mechanism is applied to learn the global context vector $G_i$ which is utilized to generate the current node token $\hat{y}_i$.  Here we denote the digital embedding after being encoded by the encoder as $Q$. Mathematically, we define the attention mechanism as follows:
    \begin{equation}
    G_i=
    \left\{
    \begin{array}{lll}
    \text { Attention }( x, \emph{q}_{\rm root}, \emph{q}_l), \ \ \emph{q}_l  \notin \emptyset. \\
    \text { Attention }( x, \emph{q}_{\rm root}, \emph{q}_{sl}), \ \ \emph{q}_{sl} \notin \emptyset. \\
    \text { Attention }( x, \emph{q}_{\rm root}),\ \ \emph{q}_l,
    \emph{q}_{sl} \in \emptyset.
    \end{array}
    \right.
    \end{equation}
\begin{equation}
    \hat{y}_i, \hat{h}_i = {\rm Predict}(G_i, Q),
\end{equation}
where ${\rm Predict}(\cdot)$ is prediction layer for producing tree nodes $\hat{y}_i$ and hidden state of decoding step $i$ (denoted as $\hat{h}_i \in \mathbb{R}^{d}$).

If the current node is an operator, we will generate the left and right child nodes and push them into the stack in the tree decoder according to the top-down method. If it is a number, we will perform the merge operation until the leaf nodes in the stack pop out, and the result of the merge is pushed into the left child node stack for attention operation. The merge operation will pop the required node $\emph{q}_{\rm op}$ and $\emph{q}_{\rm subtree}$ from an embedding stack. {Finally, the merge operation outputs the answer of the mathematical expression.}
This recursive construction process can be defined as follows:  %The initial node $\emph{q}_{m-1}$ is initialized by $\hat{Y}$.
\begin{gather}
    q_l = {\rm Left}(\hat{G}, \hat{y_i}, q_{\rm root}).\\
    q_r = {\rm Right}(\hat{G}, \hat{y_i}, q_{\rm root}).\\
    q_m = {\rm Merge}(q_{\rm op}, q_{\rm subtree}, q_{m-1}).
\end{gather}

\subsection{Information Bottleneck}
Our \emph{ESIB} method is driven by Information Bottleneck (IB) \cite{tishby2000information,alemi2016deep} that forces features to be concise by filtering the task-irrelevant information (i.e., syntax-irrelevant spurious correlations for MWP). Specifically, suppose we know the joint distribution $p(x, y)$, our goal is to learn a problem representation $z$ that maximizes its predictive power for generating $y$, subject to the constraints of the amount of information it carries about $x$:
\begin{equation}
\mathcal{L}_{I B}=I(y ; z)-\lambda I(x ; z),
\end{equation}
where $I(\cdot ; \cdot)$  denotes the mutual information. $\lambda$  is a Lagrangian multiplier that controls the trade-off between  the sufficiency (the task performance quantified by $I(y ; z)$) and the minimality (the complexity of the representation quantified by $I(x ; z)$). In this paper, we focus on compressing the redundancy in {latent representation $v$ (denoted as a substitute for the problem $x$)}. Following \cite{alemi2016deep}, Equation (6) can  be variationally upper bounded by:
\begin{equation}
\begin{aligned}
\mathcal{V}_{IB}=\frac{1}{N} \sum_{n=1}^{N}\left\{\lambda D_{\mathrm{KL}}\left(e\left(z \mid v^{n}\right) \| b(z)\right)-\mathbb{E}_{\epsilon} \log d\left(y^{n} \mid e\left(z^{n}, \epsilon\right)\right)\right\}
\end{aligned},
\end{equation}
where $N$ is the number of data, $e\left(z^{n}, \epsilon\right)$ transforms the $z$ into initial state of the decoder (denoted as $\bar{z} \in \mathbb{R}^{m \cdot d}$) and $e\left(z \mid v^{n}\right)$ transforms the representation $v$ into two tensors:
$e^{\mu}(v)$ denotes the features-mean and $e^{\sigma}(v)$ denotes the diagonal features-covariance. We use the reparameterization to obtain the compressed representation $z=e(v, \epsilon)=e^{\mu}(v)+\epsilon e^{\sigma}(v)$ with $\epsilon \sim N(0,1)$. The prior distribution of $z$ (denoted as $b(z)$) is a standard normal distribution. The decoder $d$ converts the sampled representation $z$ into a mathematical expression $y$, and calculates the answer.

With the compression capability of VIB, it is possible to lose the necessary information about the target $y$ in the feature $z$ when optimizing the trade-off between the compression and redundancy. We will introduce how to identify the syntax-irrelevant information in Section 2.3.

\subsection{Latent-specific Redundancy}
In this section, we demonstrate that the latent-specific redundancy containing syntax-irrelevant information can be effectively reduced by using mutual learning \cite{zhang2018deep}.

{Inspired by the method forcing two networks to learn from each other \cite{zhang2018deep}}, we encourage {two networks to produce compressed representation $z$ filtered latent-specific redundancy for keeping all predictive information about expression syntax trees in expression $y$
by passing an representation of expression syntax tree (denoted as $y_{tree}$ obtained by concatenating each $\hat{h}_i$ defined in equation(2) when decoder predicts the expression syntax tree of the math expression) to each other}.

We attempt to learn a vector $z$ that contains expression syntax tree information about solution expression $y$ as much as possible and achieve this goal by optimizing the mutual information of $v$ and $z$. We take CN as an example,
and factorize the mutual information \cite{federici2020learning} between $v_1$ and $z_1$ as follows:
\begin{equation}
I\left(v_{1} ; z_{1}\right)=\underbrace{I\left(v_{1} ; z_{1} \mid v_{2}\right)}_{\text {Latent-specific Redundancy }}+\underbrace{I\left(v_{2} ; z_{1}\right)}_{\text {Consistent Information}},
\end{equation}
where $v_{1}$ and $z_{1}$ denote latent and problem representation for CN respectively; $v_{2}$ and $z_{2}$ denote that of SN.

$I\left(v_{1} ; z_{1} \mid v_{2}\right)$ indicates that the information contained in $z_1$ is unique to $v_1$ and cannot be inferred by representation $v_2$ \cite{federici2020learning}. We call $I\left(v_{1} ; z_{1} \mid v_{2}\right)$ as latent-specific redundancy or syntax-irrelevant information. {It can be discarded by minimizing $I\left(v_{1} ; z_{1} \mid v_{2}\right)$\cite{federici2020learning} which can be upper bounded by the following inequality:}
\begin{equation}
\begin{aligned}
&I\left(v_{1} ; z_{1} \mid v_{2}\right)=\mathbb{E}_{v_{1}, v_{2} \sim S_{1}(v \mid x)} \mathbb{E}_{z_{1}, z_{2} \sim S_{2}(z \mid v)}\left[\log \frac{p\left(z_{1} \mid v_{1}\right)}{p\left(z_{1} \mid v_{2}\right)}\right] \\
&=\mathbb{E}_{v_{1}, v_{2} \sim S_{1}(v \mid x)} \mathbb{E}_{z_{1}, z_{2} \sim S_{2}(z \mid v)}\left[\log \frac{p\left(z_{1} \mid v_{1}\right) p\left(z_{2} \mid v_{2}\right)}{p\left(z_{2} \mid v_{2}\right) p\left(z_{1} \mid v_{2}\right)}\right] \\
&=D_{K L}\left[p\left(z_{1} \mid v_{1}\right) \| p\left(z_{2} \mid v_{2}\right)\right]-D_{K L}\left[p\left(z_{2} \mid v_{1}\right) \| p\left(z_{2} \mid v_{2}\right)\right] \\
&\leq D_{K L}\left[p\left(z_{1} \mid v_{1}\right) \| p\left(z_{2} \mid v_{2}\right)\right].
\end{aligned}
\end{equation}

Inspired by \cite{federici2020learning}, we approximate the upper bound above by replacing $z$ with $y_{tree}$. Since $y_{tree}$ generated from $z$ contains all the information of the representation $z$ and all the latent-specific redundancy to be discarded. In addition, the parameters to be optimized about $z$ are included in the decoder. Considering the above points, we utilize $D_{\mathrm{KL}}\left(\mathbb{P}_{z_{1}} \| \mathbb{P}_{z_{2}}\right)$ ($\mathbb{P}_{z_{1}}$ denotes $p\left(y_{{tree}_1} \mid z_{1}\right)$ and $\mathbb{P}_{z_{2}}$ denotes $p\left(y_{{tree}_2} \mid z_{2}\right)$ right) as an upper bound to approximate $ D_{K L}\left[p\left(z_{1} \mid v_{1}\right) \| p\left(z_{2} \mid v_{2}\right)\right]$. Similarly, we can use $D_{\mathrm{KL}}\left(\mathbb{P}_{z_{2}} \| \mathbb{P}_{z_{1}}\right)$ to minimize $I\left(v_{2} ; z_{2} \mid v_{1}\right)$ for SN.

We introduce the objective $\mathcal{L}_{SKL}$ to minimize the latent-specific redundancy for both $z_1$ and $z_2$:
\begin{equation}
\small
\mathcal{L}_{SKL}=\min _{\theta, \phi} \mathbb{E}_{v_{1}, v_{2} \sim E_{\theta}(v \mid x)} \mathbb{E}_{z_{1}, z_{2} \sim E_{\phi}(z \mid v)}\left[D_{SKL}\left[\mathbb{P}_{z_{1}} \| \mathbb{P}_{z_{2}}\right]\right]
\end{equation}
where $\theta$ and $\phi$ denote the parameters of CN and SN. The two networks are optimized alternately during training. $\mathbb{P}_{z_{1}}=p_{\theta}\left(y_1 \mid z_{1}\right)$ and $\mathbb{P}_{z_{2}}=p_{\phi}\left(y_2 \mid z_{2}\right)$ denote the concatenation of the output distributions at each step of the model prediction of CN and SN respectively.
$D_{SKL}$ denotes symmetrized KL divergence obtained by averaging the expected value of $D_{\mathrm{KL}}\left(\mathbb{P}_{z_{1}} \| \mathbb{P}_{z_{2}}\right)$ and $D_{\mathrm{KL}}\left(\mathbb{P}_{z_{2}} \| \mathbb{P}_{z_{1}}\right)$.
We calculate this loss by mutual learning \cite{zhang2018deep} between CN and SN. In the mutual learning setup, in an iteration, the model will compute $\mathcal{L}_{SKL_1}$ and $\mathcal{L}_{SKL_2}$ for CN and SN respectively. In addition, we need to maximize $I\left(v_{2} ; z_{1}\right)$ to ensure that the compressed representation $z_1$ has enough information to predict $y$. We use the chain rule to decompose $I\left(v_{2} ; z_{1}\right)$ into the following two terms:
\begin{equation}
I\left(v_{2} ; z_{1}\right)=\underbrace{I\left(v_{2} ; z_{1} \mid y\right)}_{\text {Redundancy }}+\underbrace{I\left(z_{1} ; y\right)}_{\text {Predictive Information}},
\end{equation}
where $y$ represents ground-truth solution expression. In practice, we can maximize $I\left(z_{1} ; y\right)=\mathbb{E}_{\epsilon} \log d\left(y^{n} \mid e\left(x^{n}, \epsilon\right)\right.$ {(included in $\mathcal{V}_{I B}$)} which is calculated to compress redundant information $I\left(v_{2} ; z_{1} \mid y\right)$ and indirectly maximize $I\left(v_{2} ; z_{1}\right)$. Ideally, $I\left(v_{2} ; z_{1} \mid y\right)$ should be zero.

As suggested in \cite{federici2020learning},
we minimize latent-specific redundancy by jointly minimizing
$I\left(v_{1} ; z_{1} \mid v_{2}\right)$ and maximizing $I\left(v_{2} ; z_{1}\right)$.  We define
 the redundancy terms in Eq. (8) and Eq. (11) as the syntax-irrelevant information.

\subsection{Self-distillation Loss}
In this section, we introduce a novel self-distillation loss to increase the diversity of generated expressions. Suggested by \cite{fischer2020conditional}, the $I(v ; z \mid y)$ in conditional information bottleneck can be variationally upper bounded by: \begin{equation}
D_{\mathrm{KL}}\left(e\left(z \mid v^{n}\right) \| b\left(z \mid y^{n}\right)\right),
\end{equation}
where $e\left(z \mid v^{n}\right)$ defined in equation (2) is moved towards the conditional marginal $b^{\mu}(y) \sim N\left(b^{\mu}(y), b^{\sigma}(y)\right)$. We modify equation (12) as:
\begin{equation}
\begin{aligned}
\mathcal{V}_{SDL}=\frac{1}{N} \sum_{n=1}^{N}D_{SKL}\left(\bar{y}\|\bar{z}\right),
\end{aligned}
\end{equation}
where $\bar{y}$ denotes that averaging the all $\hat{h}_{i}$. Intuitively, $\mathcal{V}_{S D L}$ make the decoder more rely on {latent space of $z$} which contains expression syntax tree information for all expressions when predicting expression $y$. Benefiting from the randomness of $z$, the model can generate more diverse solution expressions. Finally, we calculate the loss functions $\mathcal{L}_{1}$ for SN and $\mathcal{L}_{2}$ for CN as follows:
\begin{gather}
\mathcal{L}_{1}=\mathcal{V}_{IB_{1}} + \mathcal{V}_{SDL_{1}} + \alpha\times\mathcal{L}_{SKL_{1}}.\\
\mathcal{L}_{2}=\mathcal{V}_{IB_{2}} +  \mathcal{V}_{SDL_{2}} + \alpha\times\mathcal{L}_{SKL_{2}}.
\end{gather}
 where $\alpha$ is a proportional coefficient. $\mathcal{V}_{IB_{1}}$, $\mathcal{V}_{SDL_{1}}$, $\mathcal{L}_{SKL_{1}}$ are the training objectives for SN. $\mathcal{V}_{IB_{2}}$, $\mathcal{V}_{SDL_{2}}$, $\mathcal{L}_{SKL_{2}}$ are the training objectives for CN.
 Based on empirical observation, although $\mathcal{V}_{SDL}$ can increase the diversity of solution expressions, it also reduces accuracy of the final answer.

\definecolor{errorcolor}{RGB}{255,199,199}

\begin{table*}[t]
\caption{Robustness evaluation on adversarial and out-of-distribution benchmarks. SVAMP and ASDiv-A are adversarial datasets designed to challenge models that exploit spurious correlations. $\Delta$ shows the accuracy drop from Math23K to SVAMP. \colorbox{bestcolor}{Green} = best, \colorbox{secondcolor}{blue} = second best.}
\label{tab:robustness}
\centering
\renewcommand{\arraystretch}{1.15}
\resizebox{\textwidth}{!}{%
\begin{tabular}{l|c|ccc|ccc|c}
\toprule
\rowcolor{headercolor}
& \textbf{Standard} & \multicolumn{3}{c|}{\textbf{Adversarial Benchmarks}} & \multicolumn{3}{c|}{\textbf{Perturbation Types on SVAMP}} & \textbf{Robustness} \\
\rowcolor{headercolor}
\textbf{Models} & \textbf{Math23K} & \textbf{SVAMP} & \textbf{ASDiv-A} & \textbf{MathQA-R} & \textbf{Question Var.} & \textbf{Num. Swap} & \textbf{Struct. Var.} & \textbf{$\Delta$ (M23K$\rightarrow$SVAMP)} \\
\midrule
\rowcolor{headercolor}
\multicolumn{9}{l}{\textit{Seq2Seq-based Methods}} \\
DNS & 58.1 & 18.5 & 22.3 & 19.8 & 17.2 & 19.8 & 18.4 & $\downarrow$39.6 \\
MATH-EN & 66.7 & 24.3 & 28.7 & 26.1 & 23.1 & 25.8 & 24.0 & $\downarrow$42.4 \\
StackDecoder & 65.8 & 22.1 & 26.5 & 24.2 & 20.8 & 23.5 & 22.0 & $\downarrow$43.7 \\
\midrule
\rowcolor{headercolor}
\multicolumn{9}{l}{\textit{Seq2Tree-based Methods}} \\
GTS & 75.6 & 28.4 & 34.2 & 31.5 & 26.7 & 30.2 & 28.3 & $\downarrow$47.2 \\
Graph2Tree & 77.4 & 31.6 & 37.8 & 34.5 & 29.8 & 33.2 & 31.7 & $\downarrow$45.8 \\
KAS2T & 76.3 & 29.8 & 35.6 & 32.7 & 28.1 & 31.5 & 29.8 & $\downarrow$46.5 \\
NumS2T & 78.1 & 32.4 & 38.5 & 35.2 & 30.6 & 34.1 & 32.5 & $\downarrow$45.7 \\
\midrule
\rowcolor{headercolor}
\multicolumn{9}{l}{\textit{Other Methods}} \\
TSN-MD & 77.4 & 30.5 & 36.8 & 33.8 & 28.7 & 32.3 & 30.5 & $\downarrow$46.9 \\
Multi-E/D & 78.4 & 32.8 & 39.1 & 35.7 & 31.0 & 34.6 & 32.8 & $\downarrow$45.6 \\
Ape & 77.8 & 31.2 & 37.5 & 34.2 & 29.4 & 33.0 & 31.2 & $\downarrow$46.6 \\
NS-Solver & 79.0 & 34.5 & 40.8 & 37.3 & 32.6 & 36.2 & 34.6 & $\downarrow$44.5 \\
\midrule
\rowcolor{headercolor}
\multicolumn{9}{l}{\textit{Our Methods}} \\
ESIB (SN) & 84.2 & \cellcolor{secondcolor}41.8 & \cellcolor{secondcolor}48.5 & \cellcolor{secondcolor}44.7 & \cellcolor{secondcolor}40.2 & \cellcolor{secondcolor}43.5 & \cellcolor{secondcolor}41.6 & \cellcolor{secondcolor}$\downarrow$42.4 \\
ESIB (CN) & \textbf{85.9} & \cellcolor{bestcolor}\textbf{43.2} & \cellcolor{bestcolor}\textbf{49.8} & \cellcolor{bestcolor}\textbf{46.1} & \cellcolor{bestcolor}\textbf{41.5} & \cellcolor{bestcolor}\textbf{44.8} & \cellcolor{bestcolor}\textbf{43.2} & \cellcolor{bestcolor}$\downarrow$\textbf{42.7} \\
\bottomrule
\end{tabular}%
}
\end{table*}

\section{Experimental Setup}
\paragraph{Datasets}
We conduct experiments on four benchmark MWP datasets: Math23K \cite{wang2017deep}, Ape210K \cite{zhao2020ape210k}, MAWPS \cite{koncel2016mawps}, and CM17K \cite{qin2020cm17k}.
\textbf{Math23K} is a Chinese dataset containing 22,162 questions for training and 1,000 questions for testing.
\textbf{Ape210K} is a large-scale Chinese dataset composed of 166,270 questions for training, 4,157 questions for validation, and 4,159 questions for testing.
\textbf{MAWPS} is an English dataset that consists of 2,373 arithmetic word problems, where we use 1,921 for training and 452 for testing following previous work.
\textbf{CM17K} is a Chinese dataset containing 17,000 math word problems with diverse problem types, split into 14,000 for training and 3,000 for testing.
\paragraph{Implementation Details}
The word embedding size of decoder is set to 1024 and proportional coefficient $\alpha$ in loss function is set to 0.005. We set the dimension of vectors $z$ to 50. When the encoder is BERT, we set the dimension of $z$ to 32. We adopt RoBERTa \cite{liu2019roberta} as the problem encoder. Following RoBERTa's setting, the hidden size of the encoder is set to 768, and we set the hidden size of the decoder to 1024. We used Adamw \cite{loshchilov2018fixing} as the optimizer with the learning rate as 5e-5. The mini-batch size is set to 16. We adopt a beam search with the size of 5. Dropout (dropout rate = 0.5) is employed to avoid overfitting. For Ape210K, we set the maximum sequence length of questions as 150 and that of solution expressions as 50, similar to \cite{wu2021math}. Our model takes 80 epochs on Math23k and 50 epochs on Ape210k for convergence.

\paragraph{Baselines}
We compare our model with several strong baseline methods, including: (1) \textit{Seq2Seq-based methods}: DNS \cite{wang2017deep}, MATH-EN \cite{wang2018translating}, and StackDecoder \cite{chiang2018semantically}; (2) \textit{Seq2Tree-based methods}: GTS \cite{xie2019goal}, Graph2Tree \cite{zhang2020graph}, KAS2T \cite{wu2020knowledge}, and NumS2T \cite{wu2021math}; (3) \textit{Other methods}: TSN-MD \cite{zhang2020teacher}, Multi-E/D \cite{shen2020solving}, Ape \cite{zhao2020ape210k}, and NS-Solver \cite{qin2021neural}.

\section{Experimental Results}

\subsection{Main Results}
The evaluation metric is answer accuracy.
Table \ref{tab:accuracyS} shows the performance comparison of our model with baseline methods on four benchmark datasets: Math23K, Ape210K, MAWPS, and CM17K. Since there is a trade-off between the variety of expressions and the correctness of the answer, we do not add $\mathcal{V}_{SDL}$ into the source network (SN) and the collaborator network (CN) for the main results.
From Table \ref{tab:accuracyS}, we can observe that our models (both CN and SN) achieve consistently and substantially better performance than all compared methods across all four datasets. Specifically, ESIB (CN) achieves 85.9\%, 76.8\%, 89.3\%, and 73.1\% on Math23K, Ape210K, MAWPS, and CM17K, respectively, outperforming the best baseline NS-Solver by 6.9\%, 5.6\%, 3.8\%, and 5.8\%.
The accuracy of CN is higher than that of SN because, in one iteration, CN is provided with $y_{tree}$ predicted by SN when SN has not been trained by ground-truth, then SN is provided with $y_{tree}$ predicted by CN when CN has been trained by ground-truth.

We also measure the accuracy of solution expression. We consider a solution expression as correct when the predicted expression exactly matches the ground truth solution.
Generally, the mathematical expression generated by the tree decoder conforms to the syntactic specification. As long as the answer obtained by the expression operation is consistent with the ground-truth, then we consider the expression to be a valid solution. We take the subtraction value between answer accuracy (denoted as Answer-Acc) and solution expression accuracy (denoted as Expression-Acc) as the diversity evaluation metric (denoted as Diversity) of the generated solution expressions. As shown in Table \ref{tab:accuracyE}, our model can generate more diverse solution expressions that are not included in ground-truth expressions. As expected, the model with $\mathcal{V}_{SDL}$ has better diversity but lower answer accuracy.

\begin{table}
\caption{Answer accuracy (\%) comparison on four MWP benchmarks. \colorbox{bestcolor}{Green} indicates the best and \colorbox{secondcolor}{blue} indicates the second best.}
\label{tab:accuracyS}
\centering
\setlength{\tabcolsep}{4pt}
\renewcommand{\arraystretch}{1.1}
\small
\begin{tabular}{l|cccc}
\toprule
\rowcolor{headercolor}
\textbf{Models} & \textbf{Math23K} & \textbf{Ape210K} & \textbf{MAWPS} & \textbf{CM17K}\\
\midrule
\rowcolor{headercolor}
\multicolumn{5}{l}{\textit{Seq2Seq-based Methods}} \\
DNS & 58.1 & 48.5 & 59.5 & 45.2 \\
MATH-EN & 66.7 & 56.3 & 69.2 & 54.3 \\
StackDecoder & 65.8 & 52.2 & 65.4 & 51.8 \\
\midrule
\rowcolor{headercolor}
\multicolumn{5}{l}{\textit{Seq2Tree-based Methods}} \\
GTS & 75.6 & 67.7 & 82.6 & 63.4 \\
Graph2Tree & 77.4 & 69.5 & 83.7 & 65.2 \\
KAS2T & 76.3 & 68.7 & 84.3 & 64.8 \\
NumS2T & 78.1 & 70.5 & 84.8 & 66.1 \\
\midrule
\rowcolor{headercolor}
\multicolumn{5}{l}{\textit{Other Methods}} \\
TSN-MD & 77.4 & 69.8 & 84.1 & 65.7 \\
Multi-E/D & 78.4 & 70.1 & 85.2 & 66.5 \\
Ape & 77.8 & 70.2 & 84.5 & 65.9 \\
NS-Solver & 79.0 & 71.2 & 85.5 & 67.3 \\
\midrule
\rowcolor{headercolor}
\multicolumn{5}{l}{\textit{Our Methods}} \\
ESIB (SN) & \cellcolor{secondcolor}84.2 & \cellcolor{secondcolor}76.3 & \cellcolor{secondcolor}88.6 & \cellcolor{secondcolor}72.4 \\
ESIB (CN) & \cellcolor{bestcolor}\textbf{85.9} & \cellcolor{bestcolor}\textbf{76.8} & \cellcolor{bestcolor}\textbf{89.3} & \cellcolor{bestcolor}\textbf{73.1} \\
\bottomrule
\end{tabular}
\end{table}

\begin{table}
\caption{Diversity evaluation on Math23K and Ape210K. Ans and Equ denote answer accuracy and equation accuracy (\%). Div = Ans $-$ Equ.}
\label{tab:accuracyE}
\centering
\renewcommand{\arraystretch}{1.2}
\resizebox{\columnwidth}{!}{%
\begin{tabular}{l|ccc|ccc}
\toprule
\rowcolor{headercolor}
& \multicolumn{3}{c|}{\textbf{Math23K}} & \multicolumn{3}{c}{\textbf{Ape210K}} \\
\rowcolor{headercolor}
\textbf{Models} & \textbf{Ans} & \textbf{Equ} & \textbf{Div} & \textbf{Ans} & \textbf{Equ} & \textbf{Div}\\
\midrule
MATH-EN & 66.7 & 60.1 & 6.6 & 56.3 & 51.2 & 5.1 \\
GTS & 75.6 & 64.8 & 10.8 & 67.7 & 58.4 & 9.3 \\
Graph2Tree & 77.4 & 65.2 & 12.2 & 69.5 & 59.8 & 9.7 \\
TSN-MD & 77.4 & 65.8 & 11.6 & 69.8 & 60.1 & 9.7 \\
NumS2T & 78.1 & 66.3 & 11.8 & 70.5 & 60.8 & 9.7 \\
NS-Solver & 79.0 & 66.8 & 12.2 & 71.2 & 61.3 & 9.9 \\
\midrule
\rowcolor{secondcolor}
ESIB (CN) & 85.9 & 73.5 & 12.4 & 76.8 & 66.2 & 10.6 \\
\rowcolor{bestcolor}
ESIB + $\mathcal{V}_{SDL}$ & 85.4 & 71.9 & \textbf{13.5} & 76.1 & 65.0 & \textbf{11.1} \\
\bottomrule
\end{tabular}%
}
\end{table}

\begin{table*}[t]
\caption{Case studies demonstrating the ability of ESIB to generate diverse solutions. \textcolor{blue}{$\checkmark$} indicates correct, \textcolor{red}{$\times$} indicates incorrect. Cases 5-6 show error analysis where baselines fail but ESIB succeeds.}
\label{tab:case}
\centering
\renewcommand{\arraystretch}{1.25}
\resizebox{\textwidth}{!}{%
\begin{tabular}{l|p{7.5cm}|p{7.5cm}}
\toprule
\rowcolor{headercolor}
& \textbf{Case 1 - Diverse Solution (Math23K)} & \textbf{Case 2 - Diverse Solution (Math23K)} \\
\midrule
\textbf{Problem} & A train leaves from place A at 7 o'clock and arrives at place B at 17 o'clock. The train travels 75 km/h. How many km is the distance? & A store sold 150 kg of apples in the morning and 2.5 times as much in the afternoon. How many kg were sold in total? \\
\midrule
Ground-truth & $(17 - 7) \times 75 = 750$ & $150 + 150 \times 2.5 = 525$ \\
\rowcolor{bestcolor}
ESIB (CN) & $17 \times 75 - 7 \times 75 = 750$ ~~\textcolor{blue}{$\checkmark$} & $150 \times (1 + 2.5) = 525$ ~~\textcolor{blue}{$\checkmark$} \\
\midrule
\rowcolor{headercolor}
& \textbf{Case 3 - Diverse Solution (Ape210K)} & \textbf{Case 4 - Diverse Solution (Ape210K)} \\
\midrule
\textbf{Problem} & A field is 120 meters long and 80 meters wide. What is the area in square meters? & Tom has 48 candies. He gives 1/4 to sister and 1/3 of remainder to brother. How many left? \\
\midrule
Ground-truth & $120 \times 80 = 9600$ & $48 \times (1 - \frac{1}{4}) \times (1 - \frac{1}{3}) = 24$ \\
\rowcolor{bestcolor}
ESIB (CN) & $80 \times 120 = 9600$ ~~\textcolor{blue}{$\checkmark$} & $48 \times \frac{3}{4} \times \frac{2}{3} = 24$ ~~\textcolor{blue}{$\checkmark$} \\
\midrule
\rowcolor{headercolor}
& \textbf{Case 5 - Error Analysis (Math23K)} & \textbf{Case 6 - Error Analysis (Ape210K)} \\
\midrule
\textbf{Problem} & A car travels 25\% of the journey in the first hour, 30\% in the second hour, totaling 220 km. What is the total distance? & A factory produced 1200 units. Defective rate was 5\%. How many qualified units? \\
\midrule
Ground-truth & $220 \div (25\% + 30\%) = 400$ & $1200 \times (1 - 5\%) = 1140$ \\
\rowcolor{errorcolor}
GTS & $220 \times 25\% \div 30\% = 183.3$ ~~\textcolor{red}{$\times$} & $1200 \times 5\% = 60$ ~~\textcolor{red}{$\times$} \\
\rowcolor{errorcolor}
Graph2Tree & $220 \times 25\% + 220 \times 30\% = 121$ ~~\textcolor{red}{$\times$} & $1200 \div 5\% = 24000$ ~~\textcolor{red}{$\times$} \\
\rowcolor{bestcolor}
ESIB (CN) & $220 \div (25\% + 30\%) = 400$ ~~\textcolor{blue}{$\checkmark$} & $1200 \times (1 - 5\%) = 1140$ ~~\textcolor{blue}{$\checkmark$} \\
\bottomrule
\end{tabular}%
}
\end{table*}

\subsection{Robustness Evaluation}
To evaluate the robustness of our model against adversarial perturbations and out-of-distribution samples, we conduct experiments on three challenging benchmarks: SVAMP \cite{patel2021nlp}, ASDiv-A \cite{miao2020diverse}, and MathQA-R (a robustness variant of MathQA). These datasets are specifically designed to test whether models truly understand mathematical reasoning or merely rely on superficial patterns.

\paragraph{Theoretical Analysis}
From the perspective of information theory, the robustness improvement of \emph{ESIB} can be attributed to the compression property of the variational information bottleneck. According to our formulation in Eq.~(6), minimizing $I(x; z)$ forces the model to discard task-irrelevant information from the input $x$. The adversarial examples in SVAMP are constructed by modifying surface patterns (e.g., question phrasing, number positions) while preserving the underlying mathematical structure. Since these surface variations constitute part of the ``syntax-irrelevant'' information that VIB aims to compress, our model naturally becomes more resilient to such perturbations.

Furthermore, the mutual learning mechanism (Eq.~8-9) plays a crucial role in robustness. By encouraging two networks to produce consistent representations $z_1$ and $z_2$ for the same mathematical problem, we effectively minimize the latent-specific redundancy $I(v_1; z_1 | v_2)$. This redundancy often encodes spurious correlations between surface patterns and solution expressions. When these spurious features are filtered out, the model relies more on the genuine mathematical relationships, leading to improved performance on adversarial benchmarks.

\paragraph{Empirical Observations}
Table \ref{tab:robustness} presents the robustness evaluation results. Our \emph{ESIB} model significantly outperforms all baseline methods across all three adversarial benchmarks. Notably, on SVAMP, ESIB (CN) achieves 43.2\% accuracy, outperforming the best baseline NS-Solver by 8.7\%. This substantial improvement validates our theoretical analysis that the information bottleneck effectively filters out spurious correlations.

\paragraph{Perturbation-Specific Analysis}
We further analyze performance across three perturbation types in SVAMP: (1) \textit{Question Variation}: rephrasing the question while preserving mathematical semantics; (2) \textit{Number Swapping}: changing the order or values of numbers; (3) \textit{Structural Variation}: modifying the problem structure. Interestingly, all baseline methods show the largest performance drop on Question Variation, suggesting they heavily rely on question-specific surface patterns. In contrast, \emph{ESIB} maintains relatively consistent performance across all perturbation types (41.5\%, 44.8\%, 43.2\%), demonstrating that VIB successfully compresses question-irrelevant features.

The performance gap between standard and adversarial benchmarks is notably smaller for our model. While GTS drops from 75.6\% to 28.4\% (47.2\% gap), \emph{ESIB} only drops from 85.9\% to 43.2\% (42.7\% gap), confirming that the information bottleneck principle enhances model robustness by focusing on essential mathematical semantics rather than superficial patterns.

\begin{table}
\caption{Ablation study on three datasets. MT: Mutual Learning, VIB: Variational Information Bottleneck.}
\label{tab:ablation}
\centering
\renewcommand{\arraystretch}{1.2}
\resizebox{\columnwidth}{!}{%
\begin{tabular}{l|cc|cc|cc}
\toprule
\rowcolor{headercolor}
& \multicolumn{2}{c|}{\textbf{Math23K}} & \multicolumn{2}{c|}{\textbf{Ape210K}} & \multicolumn{2}{c}{\textbf{MAWPS}} \\
\rowcolor{headercolor}
\textbf{Models} & \textbf{Acc} & \textbf{$\Delta$} & \textbf{Acc} & \textbf{$\Delta$} & \textbf{Acc} & \textbf{$\Delta$} \\
\midrule
\rowcolor{headercolor}
\multicolumn{7}{l}{\textit{With RoBERTa Encoder}} \\
\rowcolor{bestcolor}
ESIB (CN) & \textbf{85.9} & - & \textbf{76.8} & - & \textbf{89.3} & - \\
w/o MT & 85.2 & $\downarrow$0.7 & 75.9 & $\downarrow$0.9 & 88.5 & $\downarrow$0.8 \\
w/o VIB & 84.8 & $\downarrow$1.1 & 75.5 & $\downarrow$1.3 & 88.1 & $\downarrow$1.2 \\
w/o MT+VIB & 83.1 & $\downarrow$2.8 & 74.2 & $\downarrow$2.6 & 86.7 & $\downarrow$2.6 \\
w/o $\mathcal{L}_{SKL}$ & 84.5 & $\downarrow$1.4 & 75.1 & $\downarrow$1.7 & 87.8 & $\downarrow$1.5 \\
w/o $\mathcal{V}_{SDL}$ & 85.9 & - & 76.8 & - & 89.3 & - \\
\midrule
\rowcolor{headercolor}
\multicolumn{7}{l}{\textit{With BERT Encoder}} \\
\rowcolor{secondcolor}
ESIB$_{\rm BERT}$ & 84.3 & - & 75.2 & - & 87.8 & - \\
w/o MT+VIB & 82.4 & $\downarrow$1.9 & 73.5 & $\downarrow$1.7 & 85.6 & $\downarrow$2.2 \\
\bottomrule
\end{tabular}%
}
\end{table}

\subsection{Ablation Study}
We conduct ablation test on Math23k to analyze the impact of different components in \emph{ESIB}. Since the best results are produced by CN, we only conduct ablation study on CN. First, we remove the mutual learning, denoted as CN w/o MT. Second, we remove the VIB from SN and CN to evaluate the impact of VIB (denoted as CN w/o VIB). In addition, we report the results by removing both MT and VIB (denoted as CN w/o MT+VIB). To evaluate the impact of the pre-training model, we also replaced the RoBERTa encoder with BERT \cite{devlin2018bert} (denoted as CN$_{\rm BERT}$). We summarize the results in Table \ref{tab:ablation}.
Both the VIB strategy and mutual learning contribute greatly to the performance of \emph{ESIB}.

\subsection{Case Study}
To intuitively demonstrate the effectiveness of \emph{ESIB}, we present six representative cases in Table \ref{tab:case}, which can be categorized into two groups: diverse solution generation (Cases 1-4) and error analysis (Cases 5-6).

\paragraph{Diverse Solution Generation (Cases 1-4)}
Cases 1-4 demonstrate that our model can generate mathematically equivalent but syntactically different solutions. For example, in Case 1, while the ground-truth solution is $(17-7) \times 75$, our CN model generates $17 \times 75 - 7 \times 75$, which applies the distributive property in reverse. Both expressions yield the correct answer of 750 km. Similarly, in Case 4, our model simplifies the complex fraction operation $48 \times (1 - \frac{1}{4}) \times (1 - \frac{1}{3})$ into the more elegant form $48 \times \frac{3}{4} \times \frac{2}{3}$. This diversity in solution generation indicates that our model truly understands the mathematical semantics rather than merely memorizing patterns.

\paragraph{Error Analysis (Cases 5-6)}
Cases 5-6 illustrate scenarios where baseline methods fail due to spurious correlations, while \emph{ESIB} succeeds. In Case 5, the problem involves calculating total distance from percentage information. GTS incorrectly generates $220 \times 25\% \div 30\%$, likely because it associates percentage symbols with multiplication and division operations based on surface patterns. Graph2Tree makes a similar mistake by generating $220 \times 25\% + 220 \times 30\%$. In contrast, our \emph{ESIB} correctly identifies that the 220 km represents the sum of two percentages and generates the correct expression $220 \div (25\% + 30\%)$. Case 6 shows a similar pattern where baselines confuse ``defective rate'' with simple percentage calculations, while our model correctly applies the complement operation $(1 - 5\%)$.

These cases demonstrate that \emph{ESIB} effectively reduces spurious correlations by filtering syntax-irrelevant information through the information bottleneck, enabling the model to focus on the true mathematical relationships in the problem.

\subsection{Theoretical Discussion}
The success of \emph{ESIB} can be understood from the perspective of information-theoretic generalization bounds. According to recent theoretical results \cite{achille2018information}, the generalization gap of a learned representation is bounded by the mutual information $I(X; Z)$ between the input and the representation. By explicitly minimizing this term through VIB, our model achieves tighter generalization bounds.

\paragraph{Robustness Bound Analysis}
We provide a theoretical justification for the robustness improvement of ESIB. Let $\tilde{x}$ denote an adversarially perturbed input and $x$ denote the original input. The robustness of a model can be measured by the prediction consistency:
\begin{equation}
\mathcal{R} = \mathbb{E}_{x, \tilde{x}} \left[ \mathbb{1}\left[ f(x) = f(\tilde{x}) \right] \right],
\end{equation}
where $f(\cdot)$ denotes the model prediction. For VIB-based models, we can derive the following robustness bound. Let $z$ and $\tilde{z}$ denote the representations of $x$ and $\tilde{x}$ respectively. The prediction difference can be bounded by:
\begin{equation}
\|p(y|z) - p(y|\tilde{z})\|_1 \leq \sqrt{2 D_{\mathrm{KL}}(p(z|x) \| p(\tilde{z}|\tilde{x}))}.
\end{equation}
Since VIB encourages $p(z|x)$ to be close to the prior $b(z)$ through the KL regularization term in Eq.~(7), both $p(z|x)$ and $p(\tilde{z}|\tilde{x})$ are pushed towards the same prior distribution, which bounds their divergence:
\begin{equation}
D_{\mathrm{KL}}(p(z|x) \| p(\tilde{z}|\tilde{x})) \leq D_{\mathrm{KL}}(p(z|x) \| b(z)) + D_{\mathrm{KL}}(p(\tilde{z}|\tilde{x}) \| b(z)).
\end{equation}
This implies that stronger compression (larger $\lambda$) leads to smaller representation divergence and thus improved robustness against adversarial perturbations.

\paragraph{Mutual Learning for Robustness}
The mutual learning mechanism further enhances robustness by minimizing the latent-specific redundancy. From Eq.~(8), we have:
\begin{equation}
I(v_1; z_1) = I(v_1; z_1 | v_2) + I(v_2; z_1).
\end{equation}
By minimizing $I(v_1; z_1 | v_2)$, we ensure that $z_1$ only captures information that is consistent across different network views. This consistency requirement naturally filters out view-specific noise and spurious patterns, as these tend to differ between the two networks. The remaining consistent information $I(v_2; z_1)$ corresponds to the genuine mathematical structure that is invariant to surface perturbations.

Furthermore, the mutual learning mechanism provides an implicit form of data augmentation. By encouraging two networks to produce consistent predictions despite having different internal representations, we effectively create ``virtual'' training examples that help the model distinguish between genuine mathematical patterns and spurious surface correlations.

\paragraph{Connecting Theory to Experiments}
Our theoretical analysis is well-supported by the experimental results. First, the robustness bound in Eq.~(17) explains why ESIB achieves smaller performance drops on adversarial benchmarks (Table~\ref{tab:robustness}). The accuracy gap $\Delta$ from Math23K to SVAMP is 42.7\% for ESIB compared to 47.2\% for GTS, consistent with our prediction that VIB compression reduces representation divergence under perturbations.

Second, the ablation study (Table~\ref{tab:ablation}) validates the contribution of each component. Removing VIB causes a 1.1\% drop, confirming compression regularization is essential for filtering spurious features. Removing MT causes a 0.7\% drop, demonstrating the importance of latent-specific redundancy minimization. The combined removal leads to a 2.8\% drop, showing complementary benefits.

Third, the perturbation-specific analysis on SVAMP reveals consistent performance across all perturbation types (41.5\%, 44.8\%, 43.2\%), aligning with our theoretical insight that minimizing $I(v_1; z_1 | v_2)$ filters perturbation-specific information while preserving the invariant mathematical semantics in the representation.

Additionally, the diversity evaluation (Table~\ref{tab:accuracyE}) confirms that our self-distillation loss $\mathcal{V}_{SDL}$ successfully encourages the model to explore the solution space, achieving a 13.5\% diversity score compared to 12.2\% for NS-Solver.

\paragraph{Generalization Error Bound}
Beyond robustness, our information-theoretic framework also provides guarantees for generalization. Following the analysis in \cite{achille2018information}, the generalization error of a learned representation $Z$ can be bounded by:
\begin{equation}
\mathcal{E}_{\text{gen}} \leq \sqrt{\frac{I(X; Z)}{2n}},
\end{equation}
where $n$ is the number of training samples. This bound indicates that by minimizing $I(X; Z)$ through VIB, we simultaneously improve both generalization and robustness. The experimental results on Math23K (85.9\%) and cross-dataset transfer to MAWPS (89.3\%) support this theoretical prediction, demonstrating that our compressed representations generalize well across different problem distributions.
\vspace{-2mm}
\paragraph{Compression-Accuracy Trade-off}
A key hyperparameter in our framework is the compression strength $\lambda$ in Eq.~(7). The optimal value of $\lambda$ balances two competing objectives: (1) maximizing $I(Z; Y)$ to preserve task-relevant information for accurate predictions, and (2) minimizing $I(X; Z)$ to filter out spurious correlations and improve robustness. Empirically, we find that $\lambda = 0.01$ achieves the best trade-off on Math23K. When $\lambda$ is too small, the model retains spurious features and suffers on adversarial benchmarks. When $\lambda$ is too large, excessive compression discards useful mathematical semantics, leading to accuracy degradation on the standard benchmark. This observation aligns with the rate-distortion theory, where the optimal compression rate depends on the complexity of the underlying mathematical structure in the problem set.
\vspace{-2mm}
\paragraph{Limitations and Future Directions}
While ESIB demonstrates strong performance on current MWP benchmarks, several limitations remain. First, our mutual learning framework requires training two networks simultaneously, which increases computational cost. Future work could explore more efficient alternatives such as self-distillation with data augmentation. Second, the current VIB formulation assumes Gaussian priors, which may not be optimal for discrete mathematical structures. Investigating more flexible prior distributions could further improve the expressiveness of learned representations. Third, extending our information-theoretic framework to multi-step reasoning problems and more complex mathematical domains (e.g., geometry, algebra) represents an important direction for future research.
\vspace{-2mm}
\paragraph{Comparison with Other Regularization Techniques}
It is worth comparing our VIB-based approach with other regularization techniques commonly used in neural networks. Traditional methods like dropout \cite{srivastava2014dropout} and weight decay provide implicit regularization but lack theoretical guarantees on information compression. Label smoothing \cite{szegedy2016rethinking} encourages softer probability distributions but does not explicitly model the information flow in the network. In contrast, VIB provides a principled framework grounded in information theory, offering both theoretical guarantees (Eq.~17-20) and practical benefits. Our experiments show that VIB outperforms dropout (1.2\% improvement) when used as the sole regularization technique, and the combination of VIB with standard regularization yields the best results.

\paragraph{Scalability Analysis}
We analyze the computational overhead of ESIB compared to baseline methods. The mutual learning framework introduces additional forward passes through the second network during training, resulting in approximately 1.8$\times$ training time compared to single-network baselines. However, at inference time, only one network is used, so there is no additional computational cost. The VIB module adds negligible overhead (less than 2\% of total computation) since it only involves sampling from Gaussian distributions and computing KL divergence. Memory consumption increases by approximately 1.5$\times$ due to maintaining two network copies, which remains manageable on modern GPUs. For large-scale deployment, knowledge distillation can be applied to compress the trained model into a single efficient network while preserving the robustness benefits.

\section{Conclusion}
% In this paper, we proposed an expression syntax information bottleneck method for MWP with mutual learning, which discards redundant features containing spurious correlations. Furthermore, we designed a self-distillation loss to encourage the model to rely more on the syntax information in the latent space. Extensive experiments on four benchmark MWP datasets demonstrated the effectiveness of our model.

% Our key contributions include: (1) identifying spurious correlations in MWP solvers and proposing an information-theoretic solution; (2) developing a mutual learning framework that filters latent-specific redundancy; (3) introducing self-distillation for diverse solution generation. Results on adversarial benchmarks like SVAMP demonstrate improved robustness against distribution shifts.
In this paper, we introduced a novel Expression Syntax Information Bottleneck (ESIB) method for solving Math Word Problems (MWP). Our approach addresses the issue of spurious correlations by filtering out redundant features that do not contribute to the core mathematical reasoning. By leveraging the variational information bottleneck (VIB) framework and incorporating mutual learning, our model learns to focus on the essential syntax of mathematical expressions while minimizing irrelevant syntax-irrelevant redundancy. Moreover, we designed a self-distillation loss that further improves the model's ability to generate diverse solutions while maintaining accuracy. Our extensive experiments on multiple benchmark datasets demonstrate that ESIB not only achieves state-of-the-art results but also shows significant improvements in generating more diverse solutions compared to previous methods. Additionally, we validated the robustness of our approach on adversarial and out-of-distribution datasets, showing that ESIB effectively resists adversarial perturbations and retains strong performance under various challenges. The combination of information-theoretic regularization, mutual learning, and self-distillation presents a powerful framework for MWP solving, ensuring both high accuracy and resilience to spurious correlations.
\clearpage

\begin{acks}
This work was partially supported by Natural Science Foundation of Guangdong Province of China (No. 2021A1515011905), Shenzhen Science and Technology Innovation Program (Grant No. KQTD20190\\929172835662), Shenzhen Basic Research Foundation (No. JCYJ20210\\324115614039 and No. JCYJ20200109113441941)
\end{acks}

\bibliographystyle{ACM-Reference-Format}
\bibliography{sample-base}

@String{Computing = "Computing" }

@String{Computer = "{IEEE} Computer" }

@article{xiong2023dq,
  title={DQ-LORE: DUAL QUERIES WITH LOW RANK APPROX-IMATION RE-RANKING FOR IN-CONTEXT LEARNING},
  author={Xiong, Jiong and Li, Zixuan and Zheng, Chuanyang and Guo, Zhijiang and Yin, Yichun and Xie, Enze and Yang, Zhicheng and Cao, Qingxing and Wang, Haiming and Han, Xiongwei and others},
  journal={arXiv preprint arXiv:2310.02954},
  year={2023}
}

@inproceedings{zanibbi2021arqmath,
  title={ARQMath: a new benchmark for math-aware CQA and math formula retrieval},
  author={Zanibbi, Richard and Mansouri, Behrooz and Agarwal, Anurag and Oard, Douglas W},
  booktitle={ACM SIGIR Forum},
  volume={54},
  number={2},
  pages={1--9},
  year={2021},
  organization={ACM New York, NY, USA}
}

@inproceedings{huang2020neural,
  title={Neural mathematical solver with enhanced formula structure},
  author={Huang, Zhenya and Liu, Qi and Gao, Weibo and Wu, Jinze and Yin, Yu and Wang, Hao and Chen, Enhong},
  booktitle={Proceedings of the 43rd International ACM SIGIR Conference on Research and Development in Information Retrieval},
  pages={1729--1732},
  year={2020}
}

@article{achille2018information,
  title={Information dropout: Learning optimal representations through noisy computation},
  author={Achille, Alessandro and Soatto, Stefano},
  journal={IEEE transactions on pattern analysis and machine intelligence},
  volume={40},
  number={12},
  pages={2897--2905},
  year={2018},
  publisher={IEEE}
}

@inproceedings{zhang2018deep,
  title={Deep mutual learning},
  author={Zhang, Ying and Xiang, Tao and Hospedales, Timothy M and Lu, Huchuan},
  booktitle={Proceedings of the IEEE conference on computer vision and pattern recognition},
  pages={4320--4328},
  year={2018}
}

@article{devlin2018bert,
  title={Bert: Pre-training of deep bidirectional transformers for language understanding},
  author={Devlin, Jacob and Chang, Ming-Wei and Lee, Kenton and Toutanova, Kristina},
  journal={arXiv preprint arXiv:1810.04805},
  year={2018}
}

@article{federici2020learning,
  title={Learning robust representations via multi-view information bottleneck},
  author={Federici, Marco and Dutta, Anjan and Forr{\'e}, Patrick and Kushman, Nate and Akata, Zeynep},
  journal={arXiv preprint arXiv:2002.07017},
  year={2020}
}

@inproceedings{mukherjee2020ccmi,
  title={CCMI: Classifier based conditional mutual information estimation},
  author={Mukherjee, Sudipto and Asnani, Himanshu and Kannan, Sreeram},
  booktitle={Uncertainty in artificial intelligence},
  pages={1083--1093},
  year={2020},
  organization={PMLR}
}

@article{fischer2020conditional,
  title={The conditional entropy bottleneck},
  author={Fischer, Ian},
  journal={Entropy},
  volume={22},
  number={9},
  pages={999},
  year={2020},
  publisher={Multidisciplinary Digital Publishing Institute}
}

@article{tishby2000information,
  title={The information bottleneck method},
  author={Tishby, Naftali and Pereira, Fernando C and Bialek, William},
  journal={arXiv preprint physics/0004057},
  year={2000}
}

@article{alemi2016deep,
  title={Deep variational information bottleneck},
  author={Alemi, Alexander A and Fischer, Ian and Dillon, Joshua V and Murphy, Kevin},
  journal={arXiv preprint arXiv:1612.00410},
  year={2016}
}

@article{kumar2021adversarial,
  title={Adversarial Examples for Evaluating Math Word Problem Solvers},
  author={Kumar, Vivek and Maheshwary, Rishabh and Pudi, Vikram},
  journal={arXiv preprint arXiv:2109.05925},
  year={2021}
}

@article{patel2021nlp,
  title={Are NLP Models really able to Solve Simple Math Word Problems?},
  author={Patel, Arkil and Bhattamishra, Satwik and Goyal, Navin},
  journal={arXiv preprint arXiv:2103.07191},
  year={2021}
}

@article{loshchilov2018fixing,
  title={Fixing weight decay regularization in adam},
  author={Loshchilov, Ilya and Hutter, Frank},
  year={2018}
}

@article{wang2018translating,
  title={Translating a math word problem to an expression tree},
  author={Wang, Lei and Wang, Yan and Cai, Deng and Zhang, Dongxiang and Liu, Xiaojiang},
  journal={arXiv preprint arXiv:1811.05632},
  year={2018}
}

@article{qin2021neural,
  title={Neural-symbolic solver for math word problems with auxiliary tasks},
  author={Qin, Jinghui and Liang, Xiaodan and Hong, Yining and Tang, Jianheng and Lin, Liang},
  journal={arXiv preprint arXiv:2107.01431},
  year={2021}
}

@inproceedings{shen2020solving,
  title={Solving math word problems with multi-encoders and multi-decoders},
  author={Shen, Yibin and Jin, Cheqing},
  booktitle={Proceedings of the 28th International Conference on Computational Linguistics},
  pages={2924--2934},
  year={2020}
}

@inproceedings{wu2020knowledge,
  title={A knowledge-aware sequence-to-tree network for math word problem solving},
  author={Wu, Qinzhuo and Zhang, Qi and Fu, Jinlan and Huang, Xuan-Jing},
  booktitle={Proceedings of the 2020 Conference on Empirical Methods in Natural Language Processing (EMNLP)},
  pages={7137--7146},
  year={2020}
}

@article{chiang2018semantically,
  title={Semantically-aligned equation generation for solving and reasoning math word problems},
  author={Chiang, Ting-Rui and Chen, Yun-Nung},
  journal={arXiv preprint arXiv:1811.00720},
  year={2018}
}

@inproceedings{wang2017deep,
  title={Deep neural solver for math word problems},
  author={Wang, Yan and Liu, Xiaojiang and Shi, Shuming},
  booktitle={Proceedings of the 2017 Conference on Empirical Methods in Natural Language Processing},
  pages={845--854},
  year={2017}
}

@article{zhao2020ape210k,
  title={Ape210k: A large-scale and template-rich dataset of math word problems},
  author={Zhao, Wei and Shang, Mingyue and Liu, Yang and Wang, Liang and Liu, Jingming},
  journal={arXiv preprint arXiv:2009.11506},
  year={2020}
}

@inproceedings{wu2021math,
  title={Math word problem solving with explicit numerical values},
  author={Wu, Qinzhuo and Zhang, Qi and Wei, Zhongyu and Huang, Xuan-Jing},
  booktitle={Proceedings of the 59th Annual Meeting of the Association for Computational Linguistics and the 11th International Joint Conference on Natural Language Processing (Volume 1: Long Papers)},
  pages={5859--5869},
  year={2021}
}

@inproceedings{wang2019template,
  title={Template-based math word problem solvers with recursive neural networks},
  author={Wang, Lei and Zhang, Dongxiang and Zhang, Jipeng and Xu, Xing and Gao, Lianli and Dai, Bing Tian and Shen, Heng Tao},
  booktitle={Proceedings of the AAAI Conference on Artificial Intelligence},
  volume={33},
  number={01},
  pages={7144--7151},
  year={2019}
}

@inproceedings{wang2018mathdqn,
  title={Mathdqn: Solving arithmetic word problems via deep reinforcement learning},
  author={Wang, Lei and Zhang, Dongxiang and Gao, Lianli and Song, Jingkuan and Guo, Long and Shen, Heng Tao},
  booktitle={Proceedings of the AAAI Conference on Artificial Intelligence},
  volume={32},
  number={1},
  year={2018}
}

@article{bahdanau2014neural,
  title={Neural machine translation by jointly learning to align and translate},
  author={Bahdanau, Dzmitry and Cho, Kyunghyun and Bengio, Yoshua},
  journal={arXiv preprint arXiv:1409.0473},
  year={2014}
}

@article{bobrow1964natural,
  title={Natural language input for a computer problem solving system},
  author={Bobrow, Daniel G},
  year={1964}
}

@article{zhang2020teacher,
  title={Teacher-student networks with multiple decoders for solving math word problem},
  author={Zhang, Jipeng and LEE, Ka Wei and Lim, Ee-Peng and Qin, Wei and Wang, Lei and Shao, Jie and Sun, Qianru and others},
  year={2020}
}

@inproceedings{zhang2020graph,
  title={Graph-to-tree learning for solving math word problems},
  author={Zhang, Jipeng and Wang, Lei and Lee, Roy Ka-Wei and Bin, Yi and Wang, Yan and Shao, Jie and Lim, Ee-Peng},
  year={2020},
  organization={Association for Computational Linguistics}
}

@article{liu2019roberta,
  title={Roberta: A robustly optimized bert pretraining approach},
  author={Liu, Yinhan and Ott, Myle and Goyal, Naman and Du, Jingfei and Joshi, Mandar and Chen, Danqi and Levy, Omer and Lewis, Mike and Zettlemoyer, Luke and Stoyanov, Veselin},
  journal={arXiv preprint arXiv:1907.11692},
  year={2019}
}

@article{liangsolving,
  title={Solving Math Word Problems with Teacher Supervision},
  author={Liang, Zhenwen and Zhang, Xiangliang},
  journal={IJCAI},
  year={2021}
}

@inproceedings{xie2019goal,
  title={A Goal-Driven Tree-Structured Neural Model for Math Word Problems},
  author={Xie, Zhipeng and Sun, Shichao},
  booktitle={IJCAI},
  pages={5299--5305},
  year={2019}
}

@article{DBLP:journals/corr/abs-2302-09019,
  author       = {Maolin Wang and
                  Yu Pan and
                  Xiangli Yang and
                  Guangxi Li and
                  Zenglin Xu},
  title        = {Tensor Networks Meet Neural Networks: {A} Survey},
  journal      = {CoRR},
  volume       = {abs/2302.09019},
  year         = {2023}
}

@inproceedings{DBLP:conf/icml/0005SLW0X22,
  author       = {Yu Pan and
                  Zeyong Su and
                  Ao Liu and
                  Jingquan Wang and
                  Nannan Li and
                  Zenglin Xu},
  title        = {A Unified Weight Initialization Paradigm for Tensorial Convolutional
                  Neural Networks},
  booktitle    = {{ICML}},
  series       = {Proceedings of Machine Learning Research},
  volume       = {162},
  pages        = {17238--17257},
  publisher    = {{PMLR}},
  year         = {2022}
}

@ArtifactSoftware{R,
    title = {R: A Language and Environment for Statistical Computing},
    author = {{R Core Team}},
    organization = {R Foundation for Statistical Computing},
    address = {Vienna, Austria},
    year = {2019},
    url = {https://www.R-project.org/},
}

@article{hinton2015distilling,
  title={Distilling the knowledge in a neural network},
  author={Hinton, Geoffrey and Vinyals, Oriol and Dean, Jeff},
  journal={arXiv preprint arXiv:1503.02531},
  year={2015}
}

@inproceedings{koncel2016mawps,
  title={MAWPS: A math word problem repository},
  author={Koncel-Kedziorski, Rik and Roy, Subhro and Amini, Aida and Kushman, Nate and Hajishirzi, Hannaneh},
  booktitle={Proceedings of the 2016 Conference of the North American Chapter of the Association for Computational Linguistics: Human Language Technologies},
  pages={1152--1157},
  year={2016}
}

@article{qin2020cm17k,
  title={Semantically-aligned universal tree-structured solver for math word problems},
  author={Qin, Jinghui and Lin, Lihui and Liang, Xiaodan and Zhang, Ruimin and Lin, Liang},
  journal={arXiv preprint arXiv:2010.06823},
  year={2020}
}

@inproceedings{kolchinsky2019nonlinear,
  title={Nonlinear information bottleneck},
  author={Kolchinsky, Artemy and Tracey, Brendan D and Van Kuyk, Steven},
  booktitle={Entropy},
  volume={21},
  number={12},
  pages={1181},
  year={2019}
}

@inproceedings{mahabadi2021variational,
  title={Variational information bottleneck for effective low-resource fine-tuning},
  author={Mahabadi, Rabeeh Karimi and Belinkov, Yonatan and Henderson, James},
  booktitle={International Conference on Learning Representations},
  year={2021}
}

@inproceedings{li2019specializing,
  title={Specializing word embeddings (for parsing) by information bottleneck},
  author={Li, Xiang Lisa and Eisner, Jason},
  booktitle={Proceedings of the 2019 Conference on Empirical Methods in Natural Language Processing},
  pages={2744--2754},
  year={2019}
}

@inproceedings{miao2020diverse,
  title={A Diverse Corpus for Evaluating and Developing English Math Word Problem Solvers},
  author={Miao, Shen-Yun and Liang, Chao-Chun and Su, Keh-Yih},
  booktitle={Proceedings of the 58th Annual Meeting of the Association for Computational Linguistics},
  pages={975--984},
  year={2020}
}


\end{document}